\documentclass[a4paper]{article}
\usepackage{RR}
\RRNo{6286}
\usepackage{hyperref}
\usepackage[english]{babel}
\usepackage[T1]{fontenc}
\usepackage{babel}
\usepackage{times}
\usepackage[latin1]{inputenc}
\usepackage{color}
\usepackage{amsmath,pifont,amssymb}
\usepackage{graphics,array,float,epsfig}
\usepackage{amsfonts,graphicx,color}
\usepackage{eclbkbox}
\usepackage{algorithm,algorithmic}
\usepackage{theorem}
\usepackage{booktabs}
\usepackage{multirow}
\def\squarebox#1{\hbox to #1{\hfill\vbox to #1{\vfill}}}
\newcommand{\qed}{\hspace*{\fill}
        \vbox{\hrule\hbox{\vrule\squarebox{.667em}\vrule}\hrule}\smallskip}

\def\argmin{\mbox{\rm{argmin}}}

\RRdate{September 2007}
\RRnbpage{26}
\RRauthor{
Aur\'elie Bugeau 
  \and
Patrick P\'erez
}
\authorhead{Bugeau \& P\'erez}
\RRtitle{Estimation \`a noyau adaptatif dans des espaces multidimensionnels h\'et\'erog\`enes}
\RRetitle{Bandwidth selection for kernel estimation in mixed multi-dimensional spaces}
\RRresume{Les m\'ethodes d'estimation \`a noyau, telles que le mean shift, ont un inconv\'enient majeur : le choix de la taille du noyau. La taille peut \^etre fixe pour l'ensemble des donn\'ees ou varier en chaque point. La s\'election de cette taille devient vraiment difficile dans le cas de donn\'ees multidimensionnelles et h\'et\'erog\`enes. Ce rapport pr\'esente une solution \`a ce probl\`eme. Il s'agit d'une extension de l'algorithme pr\'esent\'e dans \cite{Comaniciu03a}. La taille est choisie it\'erativement pour chaque type de donn\'ees, en cherchant dans un ensemble de tailles pr\'ed\'efinies celle qui donne localement les r\'esultats les plus stables. La s\'election it\'erative n\'ecessite l'introduction d'un algorithme de segmentation mean shift bas\'e sur l'estimateur dit balloon. La m\'ethode est valid\'ee dans le contexte de la segmentation d'image couleur.}
\RRabstract{Kernel estimation techniques, such as mean shift, suffer from one major drawback: the kernel bandwidth selection. The bandwidth can be fixed for all the data set or can vary at each points. Automatic bandwidth selection becomes a real challenge in case of multidimensional heterogeneous features. This paper presents a solution to this problem. It is an extension of \cite{Comaniciu03a} which was based on the fundamental property of normal distributions regarding the bias of the normalized density gradient. The selection is done iteratively for each type of features, by looking for the stability of local bandwidth estimates across a predefined range of bandwidths. A pseudo balloon mean shift filtering and partitioning are introduced. The validity of the method is demonstrated in the context of color image segmentation based on a 5-dimensional space.
}
\RRmotcle{estimateur \`a noyau, filtrage mean shift, taille de noyau}
\RRkeyword{kernel estimation, mean shift filtering, automatic bandwidth selection}
\RRprojet{Vista}
\RRtheme{\THCom \THCog}
\URRennes
\begin{document}

\makeRR
\tableofcontents
\section{Introduction}
Clustering is an important task in a wide range of applications of computer vision. Many methods exist \cite{Jain99}. Most of them rely upon some {\it a priori}. For example, for methods such as EM, the number of clusters must be known beforehand. It can be estimated by optimizing a global criterion. Other methods assume known the shape, often elliptical, of the clusters, which is often not sufficient to handle the complexity of real images. A method that does not rely on these two priors is the "mean shift" search for the modes of a kernel density estimation. The non-parametric aspect of the approach makes it very versatile to analyze arbitrary feature spaces. Hierarchical clustering methods are also non-parametric. However, they are computationally expensive and defining the stopping criterion is not simple. These reasons explain why the mean shift clustering became recently so popular in computer vision applications.

Mean shift was first introduced by Fukunaga \cite{Fukunaga75} and latter by Cheng \cite{Cheng95}. It has then been widely studied, in particular by Comaniciu \cite{Comaniciu01, Comaniciu99, Comaniciu02}. Mean shift is an iterative gradient ascent method used to locate the density modes of a cloud of points, {\it i.e.} the local maxima of its density. The estimation of the density is done through a kernel density estimation. The difficulty is to define the size of the kernel, {\it i.e.} the bandwidth matrix. The value of the bandwidth matrix highly influences the results of the mean shift clustering.

There are two types of bandwidth matrices. The first ones are fixed for the all data set. At the opposite, the variable bandwidth matrices vary along the set and capture the local characteristics of the data. Of course, the second type is more appropriate for real scenes. In fact, a fixed bandwidth affects the estimation performance by undersmoothing the tails of the density and oversmoothing the peaks. A variable bandwidth mean shift procedure has been introduced in \cite{Comaniciu02}. It is based on the sample point density estimator \cite{Hall95}. The estimation bias of this estimator decreases in comparison to the fixed-bandwidth estimators, while the covariance remains the same. The choice of a good value for the bandwidth matrix is really essential for the variable bandwidth mean shift. Indeed, when the bandwidth is not selected properly, the performance is often worse than with a fixed bandwidth. Another variable bandwidth estimator is the balloon estimator. It suffers of several drawbacks and has therefore never been used in a mean shift algorithm. However, it has been shown in \cite{Terrell92} that this estimator gives better result than the fixed bandwidth and the sample point estimators when the dimensionality of the data is higher than three. Hence, in section \ref{sec:msbal}, we will propose a new mean shift clustering algorithm based on the balloon estimator.

The bandwidth selection can be statistical analysis-based or task-oriented. Statistical analysis-based methods compute the best bandwidth by balancing the bias against the variance of the density estimate. Task-oriented methods rely on the stability of the feature space partitioning. For example, a semi parametric bandwidth selection algorithm, well adapted for variable bandwidth mean shift, has been proposed by Comaniciu in \cite{Comaniciu03a,Comaniciu01}. It works as follows. Fixed-bandwidth mean shift partitionings are run on the data for several predefined bandwidth values. Each cluster obtained is described by a normal law. Then, for each point, the clusters to which it belongs across the range of predefined bandwidths are compared. The final selected bandwidth for this point corresponds to the one, within the predefined range, that gave the most stable among these clusters. The results obtained for color segmentation were promising. However, this method has some limits. In particular, in case of a multidimensional data points composed of independent features, the bandwidth for each feature subspace should be chosen independently. Indeed, the most stable cluster is not always the same for all the feature subspaces. A solution could be to define a set of bandwidths for each domain and to partition the data using all the possible bandwidth matrices resulting from the combination of the different sets. However as the dimensions become high and/or if the sets of predefined bandwidths become large, the algorithm can become very computationally expensive.

In this paper we address the problem of data-driven bandwidth selection for multidimensional data composed of different independent features (a data point is a concatenation of different, possibly multidimensional features, thus living in a product of different feature spaces). As no statistically founded method exists for variable bandwidth and for high dimension, we concentrate on a task-oriented method, {\it i.e.} a method that relies on the stability of the feature space partitionings. Bandwidths are selected by iteratively applying the stability criteria of \cite{Comaniciu03a} for each different feature space or domain. We also introduce a new pseudo balloon mean shift which is better adapted for high dimensional feature spaces than the variable bandwidth mean shift of \cite{Comaniciu01}.

We first recall some theory on kernel density estimation (section \ref{sec:estim}) and mean shift filtering (section \ref{sec:ms}), and introduce the pseudo balloon mean shift filtering and partitioning (subsection \ref{sec:vms}). 
In section \ref{sec:algobal}, we present our algorithm for bandwidth selection algorithm in case of multivariate data and finally we show results of our algorithm for color clustering and color image segmentation (section \ref{sec:res_band}).

\section{Kernel density estimation}\label{sec:estim}
For the clarity of the paper, we start by reminding several results on fixed and variable bandwidth kernel density estimation.
\subsection{Fixed bandwidth estimator}
Given {\small$\{{\bf x}^{(i)}\}_{i=1..n}$}, {\small$n$} points in the $d$-dimensional space {\small${\mathbb{R}}^d$}, the non-parametric kernel density estimation at each point {\small${\bf x}$} is given by:
\begin{equation} \label{eq:estim1}
\begin{split}
  \widehat f({\bf x}) &= \frac{1}{n} \sum_{i=1}^{n}  K_{\bf H}({\bf x}-{\bf x}^{(i)})
\end{split}
\end{equation}
where {\small$K_{\bf H}$} is a kernel, and the bandwidth matrix, {\small${\bf H}$}, controls the size of the kernel. The shape of the kernel is constrained to be spherically symmetric. Equation \ref{eq:estim1} can also be written as:
\begin{equation}
\begin{split}
  \widehat f({\bf x}) &= \frac{1}{n|{\bf H}|^{1/2}} \sum_{i=1}^{n} K({\bf H}^{-1/2}({\bf x}-{\bf x}^{(i)}))~~.
\end{split}
\end{equation}
The theory of kernel density estimation indicates that the kernel {\small$K$} must be a bounded function with compact support satisfying:
\begin{equation}
\begin{split}
\int_{{\mathbb{R}}^d} K({\bf x}) {\rm d}{\bf x} = 1 ~,~~~~ \lim_{\|{\bf x}\|\rightarrow \infty} \|{\bf x}\|^d K(x) = 0 ~,\\
\int_{{\mathbb{R}}^d} {\bf x}K({\bf x}) {\rm d}{\bf x} = 0 ~,~~ \int_{{\mathbb{R}}^d} {\bf x}{\bf x}^TK({\bf x}) {\rm d}{\bf x} = c_k {\bf I}~,
\end{split}
\end{equation}
where {\small$c_k$} is a constant and {\small${\bf I}$} is the identity matrix.\\

In many cases fixed bandwidth kernel estimators are not a good choice to represent the data. Indeed a variable bandwidth is more appropriate to capture the local characteristics of the data. Two main variable bandwidth estimators exist. The first one allows the definition of bandwidths at the different data points and is referred to as the sample point estimator. The second one lets the bandwidth vary with the estimation points and is often referred to as the balloon estimator or nearest neighbor estimator.

\subsection{Sample point estimator}
The sample point estimator was first introduced by Breiman {\it et al.} \cite{Breiman77}. It is a mixture of similar kernels centered on data points, possibly with different bandwidths. It is defined as:
\begin{equation}\label{eq:estim3}
\begin{split}
  \widehat f({\bf x}) &= \frac{1}{n} \sum_{i=1}^{n} K_{{\bf H}({\bf x}^{(i)})}({\bf x}-{\bf x}^{(i)}) \\
&= \frac{1}{n} \sum_{i=1}^{n} \frac{1}{|{\bf H}({\bf x}^{(i)})|^{1/2}} K({\bf H}({\bf x}^{(i)})^{-1/2}({\bf x}-{\bf x}^{(i)}))~~.
\end{split}
\end{equation}
In \cite{Terrell92} the advantages and drawbacks of this estimator have been studied. The major advantages are that it is a density and that a particular choice of {\small${\bf H}(x^{(i)})$} can considerably reduce the bias \cite{Hall95}. However finding this value for multivariate data is a hard problem not yet solved. A disadvantage is that the estimate at a point may be influenced by observations very far away and not just by points nearby. In \cite{Terrell92} simulations have shown that this estimator has a very good behavior for small-to-moderate sample sizes. It deteriorates in performance compared to fixed bandwidth estimates as the sample size grows.


\subsection{Balloon estimator}
The balloon estimator was first introduced by Loftsgaarden and Quensberry \cite{Loftsgaarden65}. It is defined as:
\begin{equation}\label{eq:estim2}
\begin{split}
  \widehat f({\bf x}) &= \frac{1}{n} \sum_{i=1}^{n} K_{{\bf H}({\bf x})}({\bf x}-{\bf x}^{(i)})\\
&=  \frac{1}{n} \sum_{i=1}^{n} \frac{1}{|{\bf H}({\bf x})|^{1/2}} K({\bf H}({\bf x})^{-1/2}({\bf x}-{\bf x}^{(i)}))~~.
\end{split}
\end{equation}
This estimator allows a straightforward asymptotic analysis since it uses standard pointwise results \cite{Mack79}. On the other hand, when applied globally, the estimate typically does not integrate to 1 and thus is usually not itself a density, even when {\small$K$} is. In \cite{Terrell92} the authors have investigated the improvement that this estimator allows over fixed bandwidth kernel estimates. For data of fewer than 3 dimensions, the improvement seems to be very modest. However the balloon estimator becomes very efficient as soon as the number of dimensions becomes higher than 3.


\subsection{Quality of an estimator}\label{sec:stat_band}
The quality of an estimator depends on the closeness of {\small{$\widehat f$}} to the target density {\small$f$}. A common measure of this closeness is the mean squared error (MSE), equal to the sum of the variance and the squared bias:
\begin{equation}\label{eq:kernelprop}
\begin{split}
{\rm MSE}({\bf x}) &= E\big[(\widehat f({\bf x})-f({\bf x}))^2\big] \\
&= \mbox{var}(\widehat f({\bf x})) + \big[\mbox{Bias}(\widehat f({\bf x})) \big]^2 ~~.
\end{split}
\end{equation}
A good estimator has a small bias and a small variance. We detail in this subsection the computation of the bias and the variance for the fixed bandwidth estimator. For that purpose, we denote as {\small$\nabla f$} the gradient of function {\small$f$} and as {\small${\cal H}(f)$} the Hessian matrix of second partial derivatives. The second-order Taylor expansion of {\small$f(\bullet)$} around {\small${\bf x}$} \cite[p.94]{Wand95} is:
\begin{equation}
f({\bf{x}}+{\bf{\delta x}}) = f({\bf{x}}) + {\bf{\delta x}}^T\nabla f({\bf{x}})+ \frac{1}{2}{\bf{\delta x}}^T{\cal H}(f({\bf x})) {\bf{\delta x}}  + o({\bf{\delta x}}^T{\bf{\delta x}})~~.
\end{equation}
Applying to the fixed kernel estimator, it leads to the expectation:
\begin{equation}
\begin{split}
E\big(\widehat{f}({\bf x})\big) &= \int \frac{1}{|{\bf H}|^{1/2}}K({\bf H}^{-1/2}({\bf u}-{\bf x}))f({\bf u}) {\rm d}{\bf u} \\
&=   \big[\int K({\bf s}){\rm d}{\bf s} f({\bf{x}})+\int K({\bf s}){\bf s}^T{\rm d}{\bf s}~{{\bf H}^{1/2}}^T  \nabla f({\bf{x}})+ \\
&~~~~~~~~\int\frac{1}{2} K({\bf s}){({\bf H}^{1/2}{\bf s})}^T{\cal H}(f({\bf x})) {({\bf H}^{1/2}{\bf s})}{\rm d}{\bf s} +\int K({\bf s})o({\bf s}^T{\bf H}{\bf s}){\rm d}{\bf s}\big]~~.
\end{split}
\end{equation}
Using the kernel properties (equation \ref{eq:kernelprop}), the fact that the trace of a scalar is just the scalar, and the identity {\small$\mbox{tr}({\bf A}{\bf B})= \mbox{tr}({\bf B}{\bf A})$}, the bias of the fixed kernel estimator becomes:
\begin{equation}
\begin{split}
\mbox{Bias}(\widehat f({\bf x})) &= E\big(\widehat{f}({\bf x})\big)-f({\bf x})\\
&=c_k\mbox{tr}\big[ {{\bf H}^{1/2}}^T{\cal H}(f({\bf x})){{\bf H}^{1/2}}\big]+\int K({\bf s})o({\bf s}^T{\bf H}{\bf s}){\rm d}{\bf s}~~.
 \end{split}
\end{equation}
The variance is
\begin{equation}
\begin{split}
\mbox{var}(\widehat f({\bf x})) &= \mbox{var}\big[\frac{1}{n}\sum_{i=1}^{n}K_{\bf H}({\bf x}-{\bf x}^{(i)})\big]\\
&= \frac{1}{n|{\bf H}|^{1/2}}\big(\int (K({\bf s}))^2 {\rm d}{\bf s}f({\bf x})+\int K({\bf s})o({\bf s}^T{\bf H}{\bf s}){\rm d}{\bf s})~~.
\end{split}
\end{equation}
Several other measures exist, mainly the mean integrated squared error (MISE) or the asymptotic mean integrated squared error (AMISE). A detailed derivation of these measures can be found in \cite{Scott92} and \cite{Wand95}. As discussed in \ref{review_band}, these measures can be used to select the best value for ${\bf H}$.

\section{Mean shift partitioning} \label{sec:ms}
An appealing technique for clustering is the mean shift algorithm, which does not require to fix the (maximum) number of clusters. In this section we first remind the definition of kernel profiles and the principle of mean shift filtering and partitioning. As we are interested in variable bandwidth estimation, we give the result of \cite{Comaniciu01} in which the mean shift for the sample point estimator was developed. We then introduce a novel pseudo balloon mean shift based on the balloon estimator.

\subsection{Kernel profile}
The profile of a kernel {\small$K$} is the function {\small$k:[0,\infty) \rightarrow {\mathbb{R}}$} such that {\small$K(x) = c_kk(\|x\|^2)$}, where {\small$c_k$} is a positive normalization constant which makes {\small$K({\bf x})$} integrate to one. Using this profile the fixed bandwidth kernel density estimator can be rewritten as:
\begin{equation}
\begin{split}
\widehat f({\bf x}) &= \frac{1}{n|{\bf H}|^{1/2}} \sum_{i=1}^{n}  K({\bf H}^{-1/2}({\bf x}-{\bf x}^{(i)}))\\
     &=\frac{c_k}{n|{\bf H}|^{1/2}} \sum_{i=1}^{n}  k(\|{\bf H}^{-1/2}({\bf x}-{\bf x}^{(i)})\|^2)
\end{split}
\end{equation}
Two main kernels are used for mean shift filtering. Using a fixed bandwidth estimator, it can be shown \cite[p.139]{Scott92}\cite[p.104]{Wand95} that the AMISE measure is minimized by the Epanechnikov kernel having the profile:
\begin{equation}
k_E(x) = \left\{
  \begin{aligned}
    &1-x  ~~~~ 0 \le x \le 1\\
    &0 ~~~~~~ x > 1 ~~.\end{aligned}\right.
\end{equation}
The drawback of the Epanechnikov kernel is that it is not differentiable at the boundary of its support (for $x=1$). The second kernel is the multivariate normal one defined by the profile:
\begin{equation}
k(x) = \exp(-\frac{1}{2}x)
\end{equation}
which leads to the interesting property:
\begin{equation}
g(x) = -k'(x) = -\frac{1}{2} k(x)~~.
\end{equation}
The normalization for this profile is {\small$c_k=(2\pi)^{-d/2}$}.

\subsection{Fixed bandwidth mean shift filtering and partitioning}
Mean shift is an iterative gradient ascent method used to locate the density modes of a cloud of points, {\it i.e.} the local maxima of its density. The mean shift filtering is well described in \cite{Comaniciu02}. Here the theory is briefly reminded.

The density gradient of the fixed kernel estimator (equation \ref{eq:estim1}) is given by:
\begin{equation}
  \nabla \widehat f({\bf x}) = {\bf H}^{-1}~\widehat f({\bf x})~{\bf m}({\bf x})
\end{equation}
where {\small${\bf m}$} is the "mean shift" vector,
\begin{equation}\label{eq:ms1}
  {\bf m}({\bf x}) = \frac{\sum_{i=1}^n {\bf x}^{(i)}~g\big(\|{\bf H}^{-1/2}({\bf x}-{\bf x}^{(i)})\|^2\big)}{\sum_{i=1}^n g\big(\|{\bf H}^{-1/2}({\bf x}-{\bf x}^{(i)})\|^2\big)}~-~{\bf x} ~~~.
\end{equation}
Using exactly this displacement vector at each step guaranties convergence to the local maximum of the density \cite{Comaniciu02}. A mode seeking algorithm (algorithm \ref{algo:ms}), or mean shift filtering can be derived by iteratively computing the mean shift vector. Each computation of this vector leads to a trajectory point {\small${\bf y}^{(j)}$}. The first trajectory point {\small${\bf y}^{(1)}$} is the estimation point {\small${\bf x}$} itself while the last point {\small${\bf y}^{(t_m)}$} is the associated mode {\small${\bf z}$} . The final partition of the feature space is obtained by grouping together all the data points that converged to the same mode (algorithm \ref{algo:ms_part}).
\begin{algorithm*}[h!]
  \caption{Mean shift filtering}
  \label{algo:ms}
  Let {\small$\{{\bf x}^{(i)}\}_{i=1,\ldots ,n}$} be {\small$n$} input points in the $d$-dimensional space and {\small$\{{\bf z}^{(i)}\}_{i=1,\ldots ,n}$} their associated modes.
  For $i=1 \ldots n$
  \begin{enumerate}
  \item Initialize {\small$j=1$}, {\small${\bf y}^{(1)}= {\bf x}^{(i)}$}.
  \item Repeat
    \begin{itemize}
    \item[$\bullet$] {\small${\bf y}^{(j+1)}= {\bf y}^{(j)}+{\bf m}({\bf y}^{(j)})$} according to equation \ref{eq:ms1}.
    \item[$\bullet$] {\small$j=j+1$}.
    \end{itemize}
  Until {\small${\bf y}^{(j-1)} = {\bf y}^{(j)}$}.
  \item Assign {\small${\bf z}^{(i)}={\bf y}^{(j)}$}.
  \end{enumerate}
\end{algorithm*}
\begin{algorithm*}[h!]
  \caption{Mean shift partitioning}
  \label{algo:ms_part}
  Let {\small$\{{\bf x}^{(i)}\}_{i=1,\ldots ,n}$} be {\small$n$} input points in the $d$-dimensional space and {\small$\{{\bf z}^{(i)}\}_{i=1,\ldots ,n}$} their associated modes.
  \begin{enumerate}
  \item Run the mean shift filtering algorithm.
  \item Group together all {\small${\bf z}^{(i)}$} which are closer than {\small${\bf H}$}, {\it i.e} two modes {\small${\bf z}^{(i)}$} and {\small${\bf z}^{(j)}$} are grouped together if :
      \begin{center}$\|{\bf z}^{(i)}-{\bf z}^{(j)}\|\le \|{\bf H}\| ~~. $\end{center}
  \item Group together all {\small${\bf x}^{(i)}$} whose associated mode belongs to the same group.
  \end{enumerate}
\end{algorithm*}

Mean shift with normal kernel usually needs more iterations to converge, but yields results that are almost always better than the ones obtained with the Epanechnikov kernel. In the sequel we will only consider the multivariate normal kernel. With a {\it d}-variate Gaussian kernel, equation \ref{eq:ms1} becomes
\begin{equation}
\label{eq:ms2}
  {\bf m}({\bf x}) = \frac{\sum_{i=1}^n {\bf x}^{(i)}~\exp(-\frac{1}{2}D^2({\bf x},{\bf x}^{(i)},{\bf H}))}{\sum_{i=1}^n \exp(-\frac{1}{2}D^2({\bf x},{\bf x}^{(i)},{\bf H}))}~-~{\bf x}
\end{equation}
where
\begin{equation}
  D^2({\bf x},{\bf x}^{(i)},{\bf H})\equiv ({\bf x}-{\bf x}^{(i)})^T{\bf H}^{-1}({\bf x}-{\bf x}^{(i)})
\end{equation}
is the squared Mahalanobis distance from {\small${\bf x}$} to {\small${\bf x}^{(i)}$}. \\


\subsection{Variable bandwidth mean shift}
In the sequel we detail the mean shift using the two variable bandwidth estimators. The first version, called "variable bandwidth mean shift", is based on the sample point estimator and was introduced in \cite{Comaniciu01}. The second one is novel, since the balloon estimator has never been used in a mean shift algorithm. We will refer to the algorithm as "pseudo balloon mean shift".
\subsubsection{Variable bandwidth mean shift using sample point estimator}\label{sec:vms}
The mean shift filtering using the sample point estimator was first introduced in \cite{Comaniciu01}.
Using this estimator, equation \ref{eq:ms2} becomes:
\begin{equation}
  {\bf m}({\bf x}) = \frac{\sum_{i=1}^n |{\bf H}({\bf x}^{(i)})|^{-1/2}{\bf x}^{(i)}~\exp(-\frac{1}{2}D^2({\bf x},{\bf x}^{(i)},{\bf H}({\bf x}^{(i)})))}{\sum_{i=1}^n |{\bf H}({\bf x}^{(i)})|^{-1/2}\exp(-\frac{1}{2}D^2({\bf x},{\bf x}^{(i)},{\bf H}({\bf x}^{(i)})))}~-~{\bf x} ~~.
\end{equation}
As with fixed bandwidth kernel estimator, a mean shift filtering algorithm can be derived based on this mean shift vector. The proof of convergence of mean shift filtering using the sample point estimator can be found in \cite{Comaniciu01}.

\subsubsection{Pseudo balloon mean shift}\label{sec:msbal}
As mentioned earlier, the balloon estimator {\small$\widehat f({\bf x})$} is not always a density (does not always integrate to one), and leads to discontinuity problems. Its derivative {\small $\nabla \widehat f({\bf x})$} contains terms that depend on {\small $({\bf x}-{\bf x}^{(i)})^2$} and of {\small ${\bf H}'({\bf x})$}. Thus there is no closed-form expression for the mean shift vector. To be able to develop a mean shift filtering algorithm based on the balloon estimator, several assumptions must be made. In the context of mean shift algorithms, the bandwidth function {\small${\bf H}$} is only defined discretely at estimation points. To turn the estimator into a density and to give a closed form to the derivatives, we assume that {\small$\forall i=1\ldots n, {\bf H}'({\bf x}^{(i)})=0$}. Using the kernel profile $k$, equation \ref{eq:estim2} evaluated at data points becomes, for $i=1\ldots n$:
\begin{equation}
\widehat f({\bf x}^{(i)}) = \frac{c_k}{n}\sum_{j=1}^n \frac{1}{|{\bf H}({\bf x}^{(i)})|^{1/2}} k(\|{\bf H}({\bf x}^{(i)})^{-1/2}({\bf x}^{(j)}-{\bf x}^{(i)})\|^2)~~.
\end{equation}
Since we consider {\small${\bf H}'({\bf x}^{(i)})=0$}, its derivative is:
\begin{equation}\label{eq:bms1}
\begin{split}
\widehat \nabla f({\bf x}^{(i)}) &= \nabla \widehat f({\bf x}^{(i)})\\
&=\frac{c_k}{n|{\bf H}({\bf x}^{(i)})|^{1/2}} \sum_{j=1}^n{\bf H}({\bf x}^{(i)})^{-1}({\bf x}^{(j)}-{\bf x}^{(i)})k(\|{\bf H}({\bf x}^{(i)})^{-1/2}({\bf x}^{(j)}-{\bf x}^{(i)})\|^2)
 \\&=\frac{c_k}{n|{\bf H}({\bf x}^{(i)})|^{1/2}}{\bf H}({\bf x}^{(i)})^{-1} \sum_{j=1}^nk(\|{\bf H}({\bf x}^{(i)})^{-1/2}({\bf x}^{(j)}-{\bf x}^{(i)})\|^2)({\bf x}^{(j)}-{\bf x}^{(i)}) \\
&=\frac{1}{n} \big[\sum_{i=1}^n{\bf H}({\bf x}^{(i)})^{-1}K_{{\bf H}({\bf x}^{(i)})}({\bf x}^{(j)}-{\bf x}^{(i)}) \big] \Big[\frac{\sum_{j=1}^n{\bf x}^{(j)}K_{{\bf H}({\bf x}^{(i)})}({\bf x}^{(j)}-{\bf x}^{(i)})}{\sum_{j=1}^n K_{{\bf H}({\bf x}^{(i)})}({\bf x}^{(j)}-{\bf x}^{(i)})}-{\bf x}^{(i)}\Big]~~.
\end{split}
\end{equation}
A mean shift filtering algorithm can be derived using the last term of previous equation as the mean shift vector:
\begin{equation}\label{eq:bms2}
{\bf m}({\bf x}) = \frac{\sum_{j=1}^n{\bf x}^{(j)}K_{{\bf H}({\bf x})}({\bf x}-{\bf x}^{(j)})}{\sum_{j=1}^nK_{{\bf H}({\bf x})}({\bf x}-{\bf x}^{(j)})}-{\bf x}~~.
\end{equation}
Previous equation is only valid at the data points. Therefore, if {\small${\bf H}$} varies for each trajectory point, the mean shift filtering algorithm is not valid and its convergence can not be proved. The solution that we propose is to defined a pseudo balloon mean shift where the bandwidth varies for each estimation point but is fixed for all trajectory points. This means that the data points influencing the computation of a trajectory point are taken with the same bandwidth along all the gradient ascent trajectory. The advantage is that the estimate at a point will not be influenced by observations too far away. We then take the bandwidth {\small${\bf H}({\bf x})$} constant for all trajectory points {\small${\bf y}^{(j)}$} corresponding to the estimation point {\small${\bf x}$} (belonging to the data points). In other words, for a given starting point, the procedure amounts to a fixed bandwidth mean shift, with bandwidth depending on the starting point. We call this new mean shift pseudo balloon mean shift. The convergence of the pseudo balloon mean shift filtering if {\small${\bf H}({\bf x})^T ={\bf H}({\bf x})$} is demonstrated in appendix. The pseudo balloon mean shift partitioning algorithm is described in Algorithm \ref{algo:bms_part}. We use the minimum of the two bandwidths in step 2 to avoid the aggregation of two very distant modes.
\begin{algorithm*}[h!]
  \caption{Pseudo balloon mean shift partitioning algorithm}
  \label{algo:bms_part}
  Let {\small$\{{\bf x}^{(i)}\}_{i=1,\ldots ,n}$} be {\small$n$} input points in the $d$-dimensional space and {\small$\{{\bf z}^{(i)}\}_{i=1,\ldots ,n}$} their associated modes.
  \begin{enumerate}
  \item For {\small$i=1,\ldots ,n$}, run the mean shift filtering algorithm from ${\bf x} = {\bf x}^{(i)}$, with ${\bf H}({\bf x}) = {\bf H}({\bf x}^{(i)})$, using equation (\ref{eq:bms2}).
  \item Group together two modes {\small${\bf z}^{(i)}$} and {\small${\bf z}^{(j)}$} if:
\begin{equation*}
\|{\bf z}^{(i)}-{\bf z}^{(j)}\| \le \min(\|{\bf H}({\bf x}^{(i)})\|,\|{\bf H}({\bf x}^{(j)})\|)~~.
\end{equation*}
  \item Group together all {\small${\bf x}^{(i)}$} whose associated modes belong to the same group.
  \end{enumerate}
\end{algorithm*}

\section{Bandwidth selection for mixed feature spaces} \label{sec:algobal}
Results of mean shift filtering or partitioning always highly depend on the kernel bandwidth {\small${\bf H}$} which has to be chosen carefully. Various methods for bandwidth selection exist in literature. In particular, several statistical criteria, which generally aim at balancing the bias against the variance of an estimator, have been introduced. They are called statistical-analysis based methods, and can be applied to any method based on kernel estimation. Other techniques, only dedicated to clustering, define a criteria based on the stability of the clusters. They are called stability based methods. In subsection \ref{review_band}, we present a review of these two types of techniques. Many of these methods have proven to be very efficient. Nevertheless, none of them is really adapted to data in high dimensional heterogeneous spaces.

Therefore, in this section we propose an algorithm dedicated to the mean shift partitioning in high dimensional heterogeneous space. We assume that the $d$-dimensional input space can be decomposed as the Cartesian product of $P$ independent spaces associated to different types of information ({\it e.g.} position, color), also called feature spaces or domains, with dimension {\small$d_{\rho}, \rho = 1 \ldots P$} (where {\small$\sum_{\rho=1}^P d_{\rho} = d$}).

\subsection{Existing methods for bandwidth selection}\label{review_band}
This first subsection present a short review of existing bandwidth selection methods of both types.
\subsubsection{Statistical-analysis based methods}
Statistical methods aim at improving the quality of the kernel estimator. We remind that a good estimator is an estimator that has a small bias and a small variance. The quality is  usually evaluated by measuring the distance (MSE, MISE, AMISE...) between the estimate {\small$\widehat f$} and the target density {\small$f$}. These measures are of little practical use since they depend on the unknown density function {\small$f$} both in the variance and the squared bias term (subsection \ref{sec:stat_band}). However, the definition of the bias and the variance leads to the following property: the absolute value of the bias increases and the variance decreases as {\small${\bf H}$} increases. Therefore to minimize the mean squared error (or any other measure), we are confronted with a bias-variance trade-off.

Several good solutions can be found in literature to find the best value for {\small${\bf H}$}. In particular, we can mention the "rule of thumb" method \cite{Silverman86}, the "plug-in" rules \cite{Park90,Sheather91} and the cross validation methods \cite{Park90}\cite[p.46]{Simonoff96}. However, all these techniques have some drawbacks. The "rule of thumb" assumes that the density is Gaussian. A practical algorithm based on the plug-in rule for one dimensional data can be found in \cite{Comaniciu02}. An algorithm for the case of multivariate data is presented in \cite[p.108]{Wand95} but it is difficult to implement. Finally, cross validation methods becomes very computationally expensive for a large set of data.

For variable bandwidth, the most often used method takes the bandwidth proportional to the inverse of the square root of a first-order approximation of the local density. This is the Abramson's rule \cite{Abramson82}. Two parameters must then be tuned in advance: a proportionality constant and an initial fixed bandwidth. The proportionality constant influences a lot the result. Also, for multidimensional multimodal data, the initial fixed bandwidth is hard to determine. The application of this technique to the variable bandwidth mean shift, {\it i.e.} to the sample point estimator, has been proposed and discussed in \cite{Comaniciu01}. It leads to good results on toy examples. Evaluation of partitioning on real data is subjective, and it is hard to assert the superiority of such statistical methods. Furthermore, their application to high dimensional multimodal data is still an open problem.




\subsubsection{A stability based method}
Methods for bandwidth selection specially dedicated to clustering have also been studied. They are based on cluster validation. Many criteria determining the validity of a cluster exist \cite{Milligan85}. For example, some methods evaluate the isolation and connectivity of the clusters. Another criterion is the stability. It is based on human visual experience: real clusters should be perceivable over a wide range of scales. This criterion has been used in the scale space theory in \cite{Leung00} or in \cite{Fukunaga90} where the stability of clusters depends on the size of the clusters. An application of a stability criterion to bandwidth selection for mean shift partitioning was introduced in \cite{Comaniciu03a}. The idea is that a partition should not change when a small variation is applied to the bandwidth. \\

As no statistical methods are currently well adapted to the variable bandwidth estimation in high dimensional heterogeneous data, we decided to use the stability to validate the partitions and to find the best bandwidth at each point. The basic principle of the method that we propose is based on the one in \cite{Comaniciu03a}. The goal is to find the best bandwidth at each point within a set of predefined bandwidths. Given a set of {\small$B$} predefined matrices {\small $\{{\bf H}^{(b)},b=1 ,\ldots , B\}$}, the best bandwidth, denoted as {\small$\Upsilon({\bf x}^{(i)})$}, in this predefined set, at each point {\small${\bf x}^{(i)}$} indexed by $i$, is the one that gives the most stable clusters. The method
is composed of two main steps. \\

The first step is called bandwidth evaluation at the partition level. The mean shift partitioning is run for each of the predefined matrices. For each scale $b$ of this range, the data is divided into a certain number of clusters. For simplicity we introduce the function $c$ which, for each scale $b$, associates a data point indexed by $i$ to its corresponding cluster. If the $i$-th data point belongs to the $u$-th cluster at scale $b$, then $c(i,b) = u$. Each cluster is then represented parametrically. Indeed the stability criteria chosen in \cite{Comaniciu03a} asserts that if the clusters can be represented by normal laws, then the best cluster is the one for which the normal law is the most stable. If few points are added to the partition or if some are left apart, the distribution of the cluster should not change. The assumption of normality seems reasonable in a small neighborhood of a point, this neighborhood being found by the partitioning. Each cluster indexed by {\small$u$} at scale {\small$b$} is then represented parametrically by a normal law {\small${\cal N}({\bf \mu}_u^{(b)},{\bf \Sigma}_u^{(b)})$}. Let {\small${\cal C}_u^{(b)}$} be the set of indices of points belonging to a cluster {\small$u$} at scale $b$, {\small${\cal C}_u^{(b)}=\{i / c(i,b)=u\}$}. The mean ${\bf \mu}_{u}^{(b)}$ corresponding to cluster {\small$u$} at scale {\small$b$} is defined as:
\begin{equation}\label{eq:mean3}
  {\bf \mu}_{u}^{(b)} = \frac{1}{|{\cal C}_u^{(b)}|}\sum_{i\in {\cal C}_u^{(b)}} {\bf x}^{(i)}~~,
\end{equation}
and the empirical covariance {\small${\bf \Sigma}_{u}^{(b)}$} as:
\begin{equation}\label{eq:sigma3}
  {\bf \Sigma}_{u}^{(b)} = \frac{1}{|{\cal C}_u^{(b)}|}\sum_{i\in {\cal C}_u^{(b)}} ({\bf x}^{(i)} - {\bf \mu}_{u}^{(b)})({\bf x}^{(i)} - {\bf \mu}_{u}^{(b)})^T ~~.
\end{equation}
These expectation and covariance estimates are easily corrupted by non-Gaussian tails which might occur. In \cite{Comaniciu03a}, other formula have been established to solve this problem. However, the proposed covariance does not seem reliable since it can be negative. Therefore, in the sequel, we will consider that all the means and covariances are computed with the traditional definitions. This choice gave satisfactory results for all the tests we have run but we believe that further work should concentrate on finding a better way to compute the covariance based on existing techniques for robust covariance matrix estimation \cite{Pena01,Wang02}.

After building all the normal laws, each point is associated at each scale to the law of the cluster it belongs to. The point indexed by $i$ is associated for scale $b$ to the distribution {\small$p_i^{(b)}={\cal N}({\bf \mu}_{c(i,b)}^{(b)},{\bf \Sigma}_{c(i,b)}^{(b)})$}. \\

The second step evaluates for each point the clusters to which this point was associated and finds the most stable one. This second step is called bandwidth evaluation at the data level. It mainly consists in the comparison of the clusters, through the comparison of the normal laws. Several divergence measures between multiple probability distributions have been studied in literature. In \cite{Comaniciu03a}, the authors use the Jensen-Shannon divergence to compare the distributions. Given {\small$r$} {\small$d$}-variate normal distributions {\small$p_{j}~,~j=1,\ldots , r$}, defined by their mean {\small${\bf \mu}_j$} and covariance {\small${\bf \Sigma}_j$}, the Jensen-Shannon divergence is defined as:
\begin{equation}\label{eq:JS0}
JS(p_1 \ldots p_r) =\frac{1}{2} {\rm log} \frac{|\frac{1}{r}\sum_{j=1}^{r} {\bf \Sigma}_j|}{\sqrt[r]{\prod_{j=1}^{r}|{\bf \Sigma}_j|}}+\frac{1}{2}\sum_{j=1}^{r} ({\bf \mu}_j-\frac{1}{r}\sum_{j=1}^{r} {\bf \mu}_j)^T(\sum_{j=1}^{r} {\bf \Sigma}_j)^{-1} ({\bf \mu}_j-\frac{1}{r}\sum_{j=1}^{r} {\bf \mu}_j) ~~.
\end{equation}
The comparison is done between three neighboring scales ($r=3$) and for each domain independently, the distributions being {\small$p_{i}^{b-1}$}, {\small$p_{i}^b$} and {\small$p_{i}^{b+1}$}. The best scale $b^*=\argmin_b JS(p_i^{(b-1)},p_i^{(b)},p_i^{(b+1)})$ is the one for which the Jensen-Shannon divergence,
\begin{equation}\label{eq:JS}
\begin{split}
JS&(p_{i}^{(b-1)}, p_{i}^{(b)},p_{i}^{(b+1)}) =\frac{1}{2} {\rm log} \frac{|\frac{1}{3}\sum_{j=b-1}^{b+1} {\bf \Sigma}_{c(i,b)}^{(j)}|}{\sqrt[3]{\prod_{j=b-1}^{b+1}|{\bf \Sigma}_{c(i,b)}^{(j)}|}}+\\
&\frac{1}{2}\sum_{j=b-1}^{b+1} ({\bf \mu}_{c(i,b)}^{(j)}-\frac{1}{3}\sum_{j=b-1}^{b+1} {\bf \mu}_{c(i,b)}^{(j)})^T(\sum_{j=b-1}^{b+1} {\bf \Sigma}_{c(i,b)}^{(j)})^{-1} ({\bf \mu}_{c(i,b)}^{(j)}-\frac{1}{3}\sum_{j=b-1}^{b+1} {\bf \mu}_{c(i,b)}^{(j)}) ~~,
\end{split}
\end{equation}
is minimized. The final best bandwidth for the point {\small${\bf x}^{(i)}$} is the predefined matrix that gave the most stable cluster: {\small${\bf \Upsilon}({\bf x}^{(i)}) = {\bf H}^{(b^*)}$}. This is in the contrast with the original method in \cite{Comaniciu03a}, where {\small${\bf \Upsilon}({\bf x}^{(i)}) = {\bf \Sigma}_{c(i,b^*)}^{(b^*)}$}. The latter choice does not guarantee that the estimated bandwidth lies between the extremal bandwidths of the selected range (${\bf H}^{(1)}$ and ${\bf H}^{(B)}$ when they are sorted).\\ 

\begin{figure}[h!]
\fbox{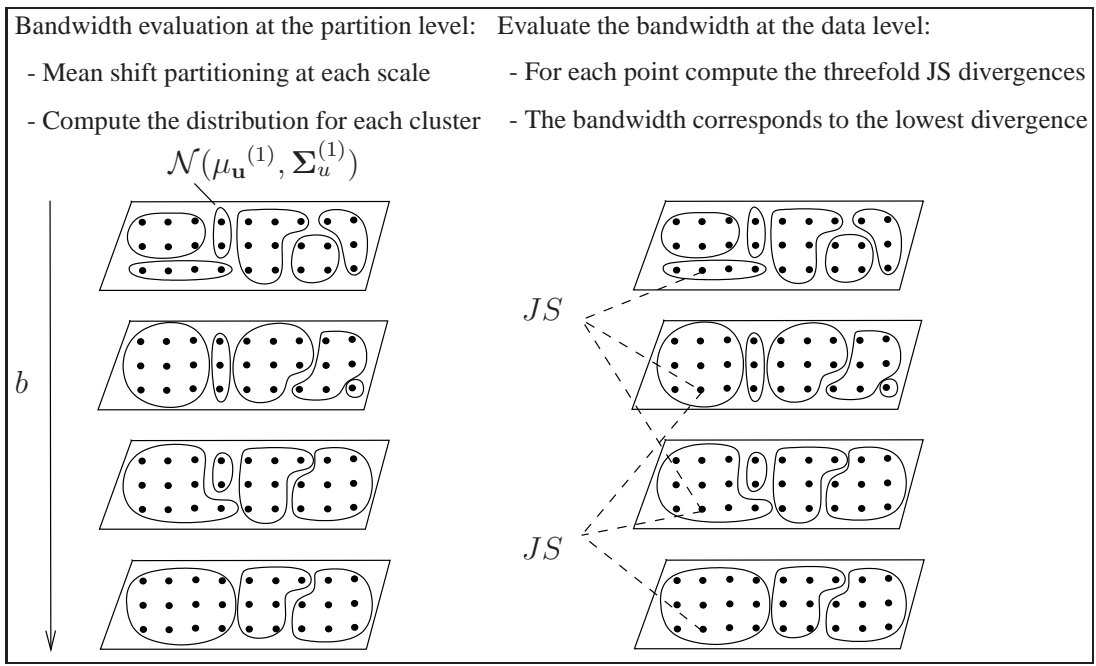\hspace{0.4cm}}
\caption{Scheme of an iteration of our algorithm.}
\label{fig:coman}
\end{figure}

These two steps are described in Figure \ref{fig:coman}. The final partition of the data is obtained by rerunning a variable or a pseudo balloon mean shift partitioning using the selected matrices. Unfortunately, this algorithm is limited to data composed of one feature space. Therefore, in next subsection we propose an iterative algorithm, based on the previous method, that handles the heterogeneity of the data.
\vspace{0.5cm}

\subsection{Handling heterogeneity: iterative selection}
For high dimensional heterogeneous data the set of predefined bandwidths can become large. Indeed, if the different domains of multidimensional data are independent, the bandwidth for each feature space should be chosen independently. The most stable cluster is not obtained for the same scale for all the domains. A solution could be to define a set of bandwidths for each feature space and to partition the data using all the possible bandwidth matrices resulting from the combination of the different sets. However as the dimensions become high and/or the sets of predefined bandwidths become larges, the algorithm can become computationally very expensive: if we have a range of {\small  $B_{\rho}$} analysis bandwidths for each feature space $\rho$, the mean shift partitioning has to be run {\small $\prod_{\rho=1}^{P}B_{\rho}$} times to take into account every possibility. Thus the algorithm is not adapted for spaces composed of several independent components. The solution is then to find the best bandwidth iteratively for each feature space, so that the mean shift partitioning is run only {\small $\sum_{\rho=1}^{P}B_{\rho}$} times.

Suppose that we are trying to select the best bandwidth at each data point for the first feature space. We fix temporary matrices {\small ${\bf {\tilde H}}_{\rho}, \rho=2,\ldots , P$} for each of the other feature spaces. These matrices are constant for all scales and equal to the mean over all the {\small $B_{\rho}$} possible matrices:
\begin{equation}
\label{eq:htemp0}
  {\bf {\tilde H}}_{\rho} = \frac{1}{B_{\rho}}\sum_{b=1}^{B_{\rho}} {\bf H}_{\rho}^{(b)} , \rho>1~~.
\end{equation}
The bandwidth selection algorithm previously defined (Figure \ref{fig:coman}) is run for the matrix range
\begin{equation*}
    \{{\bf {\tilde H}}^{(b)} = {\rm diag}[{\bf H}_{1}^{(b)},{\bf {\tilde H}}_{2},\ldots,{\bf {\tilde H}}_{P}], b=1,\ldots , B_1\}
  \end{equation*}
and finds the best bandwidth {\small${\bf \Upsilon}_1({\bf x}^{(i)})$} for each point {\small${\bf x}^{(i)}$}. The same procedure is then run for every other feature space. The difference is that for the feature spaces that have already been studied the bandwidth matrix is not constant anymore:
\begin{equation}
  \label{eq:htemp1}
  {\bf {\tilde H}}^{(b)}({\bf x}^{(i)}) = {\rm diag}[{\bf \Upsilon}_{1}({\bf x}^{(i)}),\ldots,{\bf \Upsilon}_{\rho-1}({\bf x}^{(i)}),{\bf H}_{\rho}^{(b)},{\bf {\tilde H}}_{\rho+1}\ldots{\bf {\tilde H}}_{P}]~~.
\end{equation}
A variable bandwidth mean shift must then be used. As the dimension of the data is higher than 3, we prefer the balloon based mean shift partitioning, but the sample point estimator could be used as well within the same procedure.\\

\subsection{Bandwidth selection final algorithm}
The proposed iterative algorithm solves the bandwidth selection for high dimensional heterogeneous data problem. Each feature space is processed successively in two stages. The first stage consists in partitioning the data for each scale and building a parametric representation of each cluster. The second stage selects for each data point the most stable cluster which finally leads to the best bandwidth. The final algorithm is presented in algorithm \ref{algo:Iterativeestimations}.
\begin{algorithm*}[ht]
  \caption{Iterative estimation of bandwidths}
  \label{algo:Iterativeestimations}
 Given a set of {\small$B_{\rho}$} predefined bandwidths {\small $\{{\bf H}_{\rho}^{(b)},b=1 \ldots B\}$} for each feature space {\small$\rho$}. The bandwidth selection is as follows. \\
  For $\rho = 1,\ldots , P$
  \begin{itemize}
  \item[$\bullet$] Evaluate the bandwidth at the partition level: For all {\small$b=1,\ldots , B$}
    \begin{enumerate}
    \item For all {\small$\rho'=\rho+1,\ldots , P$}, compute {\small${\bf {\tilde H}}_{\rho'}$}:
      {\small\begin{equation}
        {\bf {\tilde H}}_{\rho'} =            \frac{1}{B_{\rho'}}\sum_{b=1}^{B_{\rho'}} {\bf H}_{\rho'}^{(b)} ~~.
      \end{equation}}
    \item Define, for $i=1,\ldots,n$,
     \begin{center}{\small$\{{\bf {\tilde H}}^{(b)}({\bf x}^{(i)}) = {\rm diag}[{\bf \Upsilon}_{1}({\bf x}^{(i)}),\ldots,{\bf \Upsilon}_{\rho-1}({\bf x}^{(i)}),{\bf H}_{\rho}^{(b)},{\bf {\tilde H}}_{\rho+1}\ldots{\bf {\tilde H}}_{P}] , b=1,\ldots , B_{\rho}\}$}.\end{center}
    \item Partition the data using the balloon mean shift partitioning. The result is {\small$n^{(b)}$} clusters denoted as {\small${\cal C}_u^{(b)}, u=1\ldots n^{(b)}$}. Introduce the function $c$ that associates a point indexed by $i$ to its cluster $u$: {\small$c(i,b) = u$}.
    \item Compute the parametric representation {\small${\cal N}({\bf \mu}_u^{(b)},{\bf \Sigma}_u^{(b)})$} of each partition using:
        {\small\begin{equation}
  {\bf \mu}_{u}^{(b)} = \frac{1}{|{\cal C}_u^{(b)}|}\sum_{i\in {\cal C}_u^{(b)}} {\bf x}^{(i)} = \left[\begin{array}{c}{\bf \mu}_{u,1}^{(b)}\\\vdots \\{\bf \mu}_{u,P}^{(b)}\end{array}\right]~~,
\end{equation}}
and
{\small\begin{equation}
  {\bf \Sigma}_{u}^{(b)} = \frac{1}{|{\cal C}_u^{(b)}|}\sum_{i\in {\cal C}_u^{(b)}} ({\bf x}^{(i)} - {\bf \mu}_{u}^{(b)})({\bf x}^{(i)} - {\bf \mu}_{u}^{(b)})^T = {\rm diag}[{\bf \Sigma}_{u,1}^{(b)},\ldots ,{\bf \Sigma}_{u,P}^{(b)}]~~.
\end{equation}}
    \item Associate to each point {\small${\bf x}^{(i)}$} the mean {\small${\bf \mu}_{c(i,b),\rho}^{(b)}$} and covariance {\small${\bf \Sigma}_{c(i,b),\rho}^{(b)}$} of the cluster it belongs to and the corresponding normal distribution {\small$p^{(b)}_{i,\rho}$}.
    \end{enumerate}
  \item[$\bullet$] Evaluate the bandwidth at the data level: For each point {\small${\bf x}^{(i)}$}
    \begin{enumerate}
    \item Select the scale $b^*$ giving the most stable normal distribution by solving:
      {\small\begin{equation}
        b^* = \argmin_{r=2,\ldots , B-1} {\rm JS}(p_{i,\rho}^{(r-1)},p_{i,\rho}^{(r)},p_{i,\rho}^{(r+1)})
      \end{equation}}
      where ${\rm JS}$ is the Jensen-Shanon divergence defined by equation (\ref{eq:JS}).
    \item The best bandwidth {\small$\Upsilon_{\rho}({\bf x}^{(i)})$} is {\small${\bf H}_{\rho}^{(b^*)}$}.
    \end{enumerate}
  \end{itemize}
\end{algorithm*}

\section{Experimental results}\label{sec:res_band}
This section presents some results of our method on color image segmentation. The final partition of the data is obtained by applying a last time the pseudo balloon mean shift partitioning with the selected variable bandwidths. The data set is the set of all pixels of the image. To each data point is associated a 5-dimensional feature vector: two dimensions for the position and three for the color. We here consider the independency of all the dimensions, {\it i.e.} 5 features spaces each composed of one dimension. The order in which the dimensions are processed by our algorithm is the following: x coordinate, y coordinate, red, green and blue channels. In the final subsection we discuss the influence of the order in which the feature spaces are processed. For all the results presented in this section the same predefined bandwidths are used. For all the feature spaces, we used 9 predefined bandwidths in the range of 10-30. Of course this range is large and it would be better to adapt it to each image, for example by using some information on the noise as in \cite{CVPR07}, but it is sufficient to validate our algorithm. The color of a pixel in the segmented images corresponds to the color of its associated mode.

The two novel features of our algorithm are successively validated with comparisons to other methods. First we validate the iterative bandwidth selection in independent feature spaces by comparing our algorithm with the same method in which the bandwidths evolve jointly in the different feature spaces. We then compare the variable mean shift based on the sample point estimator and the pseudo balloon mean shift. 
Several results for each of these two points are shown.

\subsection{Validation of the iterative bandwidth selection}
We start by validating the iterative selection on several examples. The comparison is done between our algorithm with five feature spaces and our algorithm with a single five-dimensional feature space. In the last case, all dimensions are consider dependent and the same scale is finally selected for all dimensions.

\begin{figure}[h!]
\begin{center}
\begin{tabular}{ccc}
\hspace{-0.2cm}\includegraphics[scale=0.85]{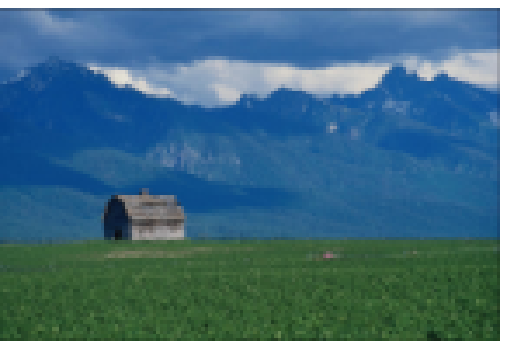}&\includegraphics[scale=0.85]{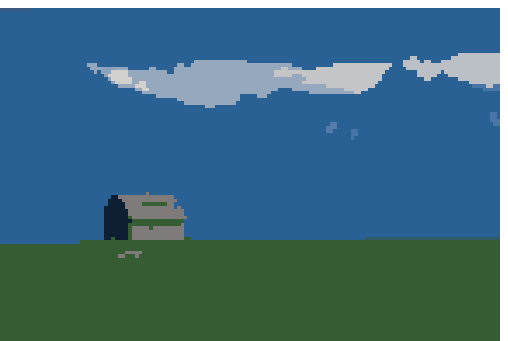}&
\includegraphics[scale=0.85]{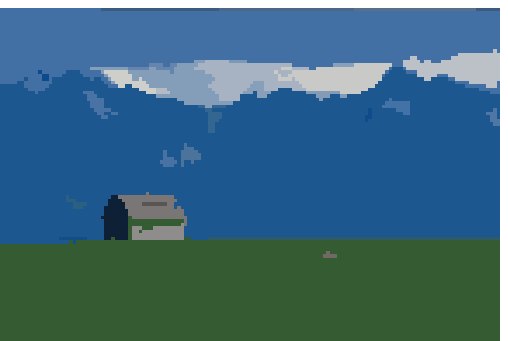}\\
\hspace{-0.2cm}\mbox{a)} & \mbox{b)} & \mbox{c)}\vspace{-0.2cm}
\end{tabular}
\caption{Validation of the iterative selection on the outdoor image.  a) Original image; b) Non iterative bandwidth selection; c) Iterative bandwidth selection.}
\label{fig:chalet1}
\end{center}
\end{figure}

\begin{figure}[ht!]
 \begin{center}
 \begin{tabular}{cccc}
 \hspace{0.5cm}\includegraphics[scale=0.6]{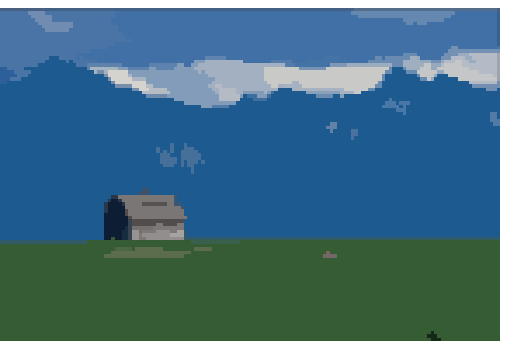}&
\includegraphics[scale=0.6]{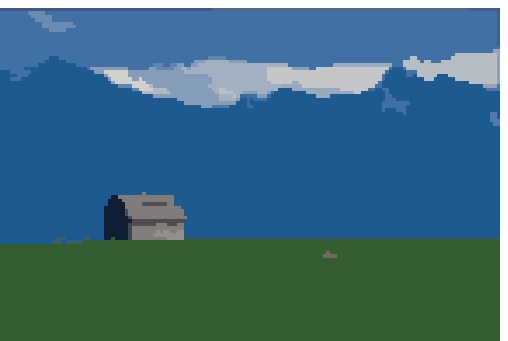}&
\includegraphics[scale=0.6]{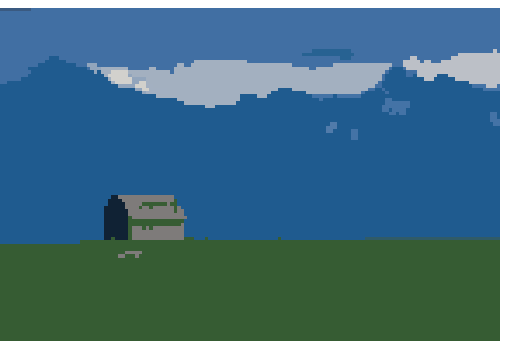}&
\includegraphics[scale=0.6]{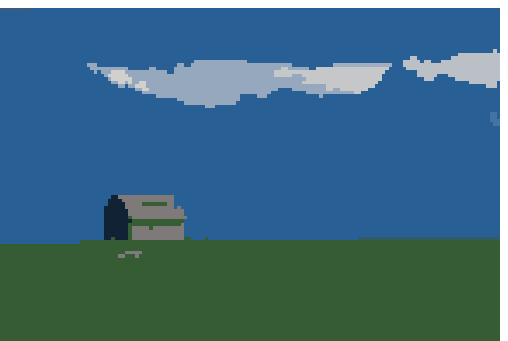}\vspace{-0.2cm}\\
\hspace{0.5cm}\mbox{$b=2$} & \mbox{$b=4$} & \mbox{$b=6$} &  \mbox{$b=8$}\vspace{-0.2cm} \\
\end{tabular}
\caption{Evolution through scales of the partitionings for our non iterative algorithm.}
\label{fig:evol3}
\vspace{0.5cm}
\begin{tabular}{ccccc}
\mbox{$\rho=1$} & & & & \vspace{-0.6cm}\\
&\hspace{-0.8cm}\multirow{6}{*}{\includegraphics[scale=0.6]{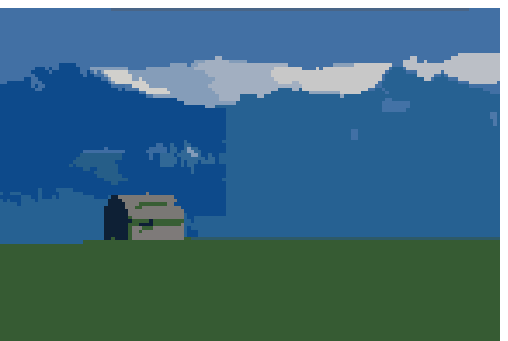}}
&\multirow{6}{*}{\includegraphics[scale=0.6]{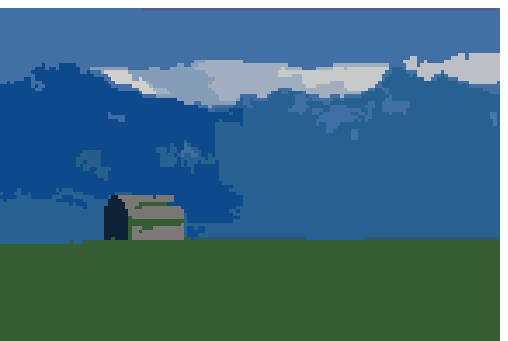}}&
\multirow{6}{*}{\includegraphics[scale=0.6]{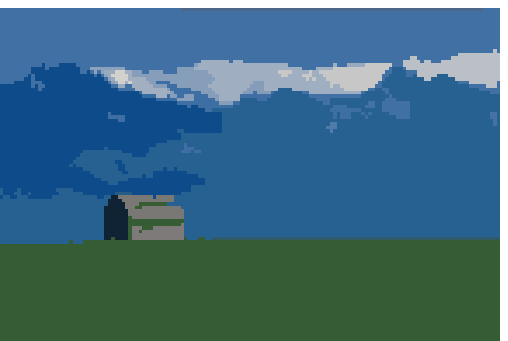}}&
\multirow{6}{*}{\includegraphics[scale=0.6]{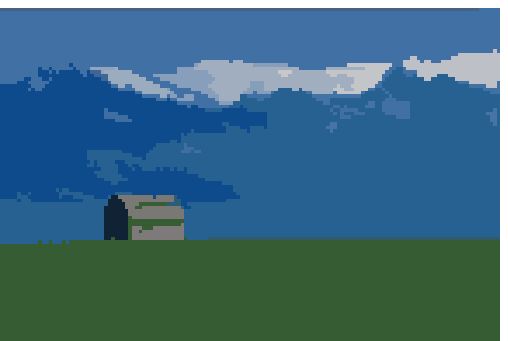}}\\
& & & & \\
& & & & \\
& & & & \\
& & & & \\
& & & & \vspace{-0.8cm}\\
\mbox{$\rho=2$} & & & & \vspace{-0.6cm}\\
&\hspace{-0.8cm}\multirow{6}{*}{\includegraphics[scale=0.6]{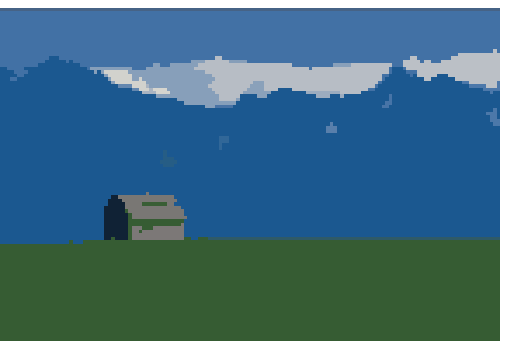}}&
\multirow{6}{*}{\includegraphics[scale=0.6]{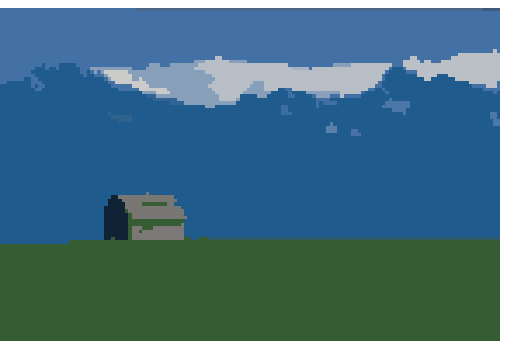}}&
\multirow{6}{*}{\includegraphics[scale=0.6]{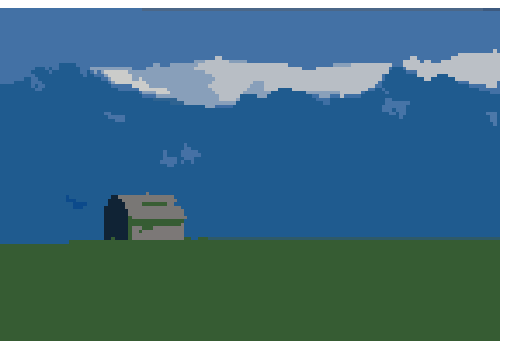}}&\multirow{6}{*}{\includegraphics[scale=0.6]{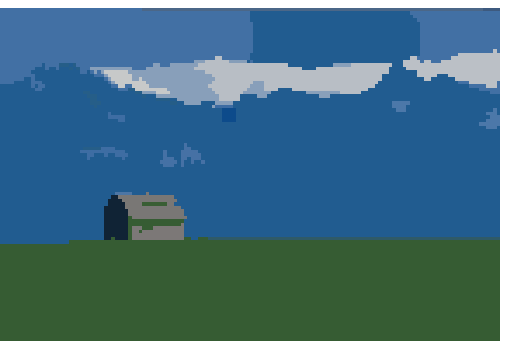}}\\
& & & & \\
& & & & \\
& & & & \\
& & & & \\
& & & &\vspace{-0.8cm} \\
\mbox{$\rho=3$} & & & & \vspace{-0.6cm}\\
&\hspace{-0.8cm}\multirow{6}{*}{\includegraphics[scale=0.6]{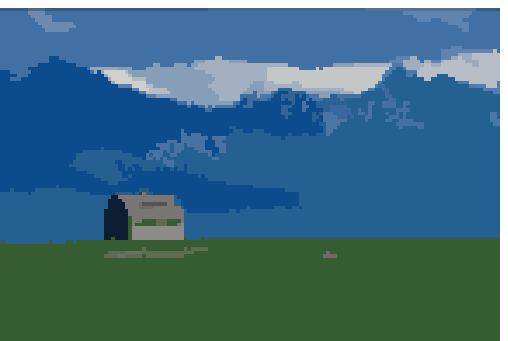}}&\multirow{6}{*}{\includegraphics[scale=0.6]{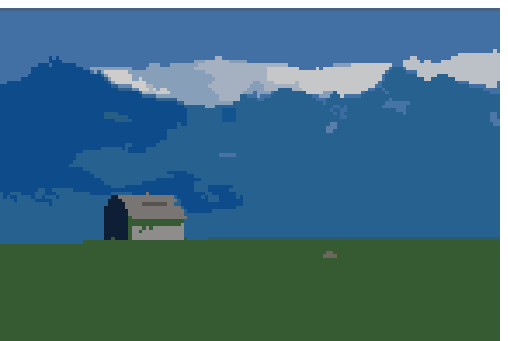}}&\multirow{6}{*}{\includegraphics[scale=0.6]{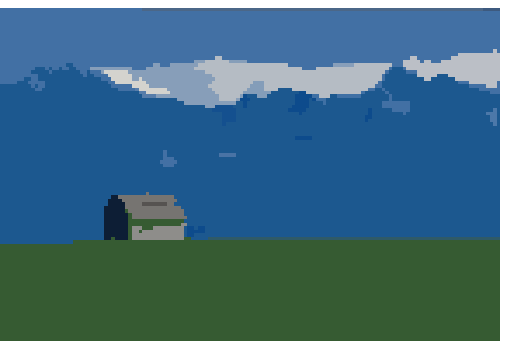}}&\multirow{6}{*}{\includegraphics[scale=0.6]{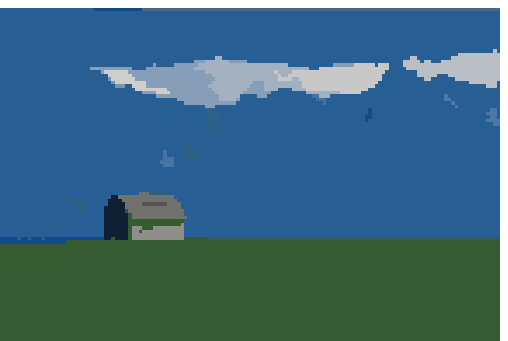}}\\
& & & & \\
& & & & \\
& & & & \\
& & & & \\
& & & &\vspace{-0.8cm} \\
\mbox{$\rho=4$} & & & & \vspace{-0.6cm}\\
&\hspace{-0.8cm}\multirow{6}{*}{\includegraphics[scale=0.6]{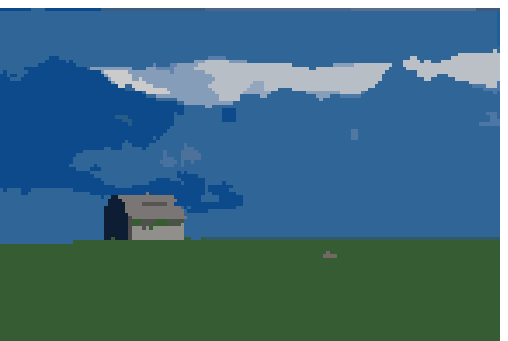}}&\multirow{6}{*}{\includegraphics[scale=0.6]{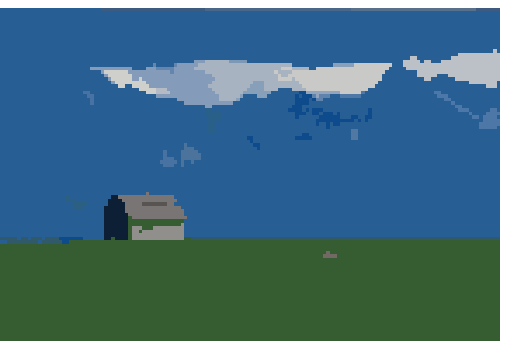}}&\multirow{6}{*}{\includegraphics[scale=0.6]{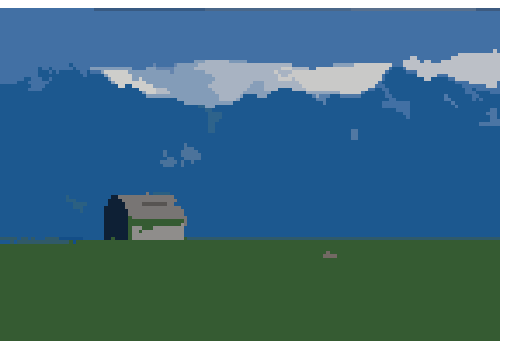}}&\multirow{6}{*}{\includegraphics[scale=0.6]{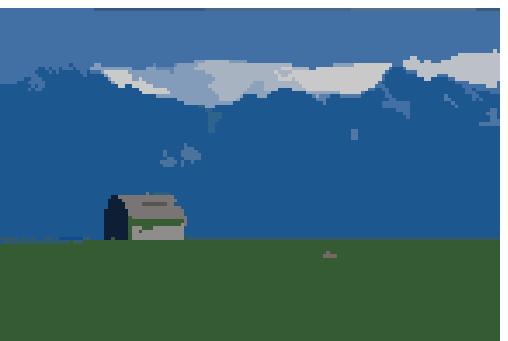}}\\
& & & & \\
& & & & \\
& & & & \\
& & & & \\
& & & &\vspace{-0.8cm} \\
\mbox{$\rho=5$} & & & & \vspace{-0.6cm}\\
&\hspace{-0.8cm}\multirow{6}{*}{\includegraphics[scale=0.6]{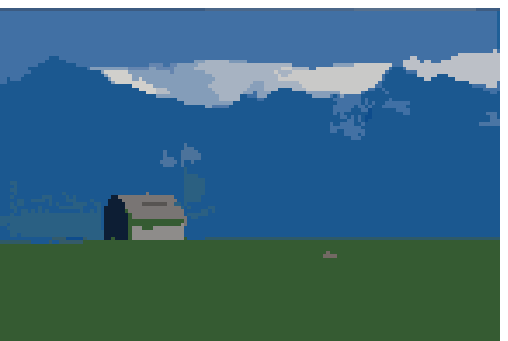}}&\multirow{6}{*}{\includegraphics[scale=0.6]{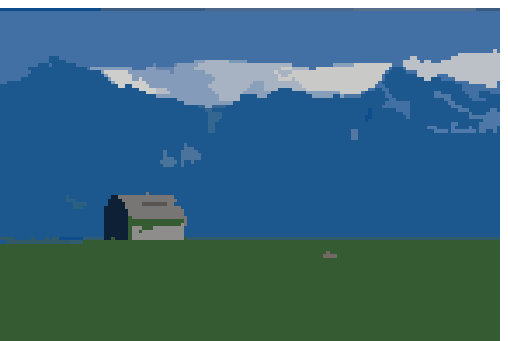}}&\multirow{6}{*}{\includegraphics[scale=0.6]{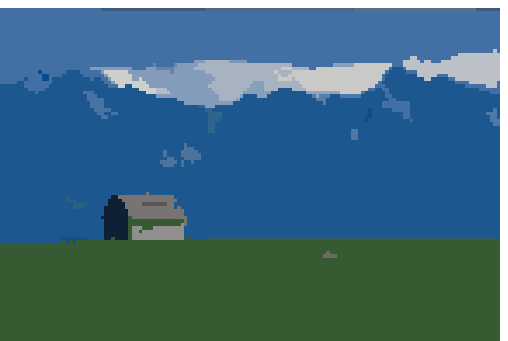}}&\multirow{6}{*}{\includegraphics[scale=0.6]{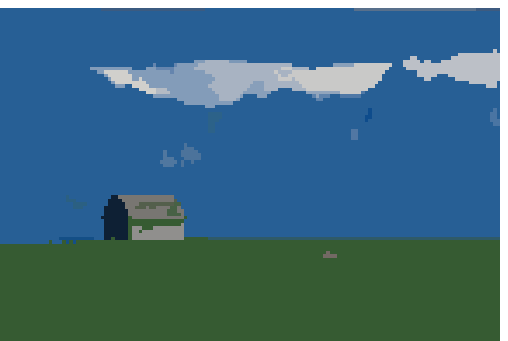}}\\
& & & & \\
& & & & \\
& & & & \\
& & & & \\
& & & &\vspace{-0.8cm} \\
&\hspace{-0.8cm}\mbox{$b=2$} & \mbox{$b=4$} & \mbox{$b=6$} &  \mbox{$b=8$} \\
\end{tabular}
\caption{Evolution through scales and feature spaces of partitionings with our algorithm.}
\label{fig:evol4}
\end{center}
\end{figure}

The first results are presented on an outdoor image. Figure \ref{fig:chalet1} shows the final partitioning for the non iterative selection Figure \ref{fig:chalet1}(b)) and the iterative selection Figure \ref{fig:chalet1} (c)). With the non-iterative algorithm, 21 clusters were found, while the iterative method gave 31 clusters. At the end of the segmentation the sky and the mountains are merged together with the non iterative algorithm. Differences are also visible on the mountains, in which the iterative method gave more clusters.
In figure \ref{fig:evol3} the evolution through scales of the mean shift partitioning that corresponds to the first step of the non iterative algorithm (``evaluate the bandwidth at the partition level'') is shown. The evolution for our iterative algorithm is presented in figure \ref{fig:evol4}. This time the evolution is shown through scales and through feature spaces. Because bandwidths evolve jointly in the different feature spaces with the non-iterative algorithm, many details are rapidly lost. Our algorithm allows more stability between two consecutive scales.

\begin{figure}[h!]
\begin{center}
\begin{tabular}{ccc}
\hspace{-0.2cm}\includegraphics[scale=0.6]{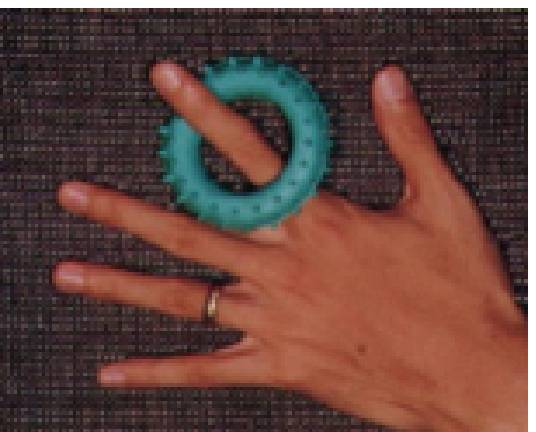}&\includegraphics[scale=0.6]{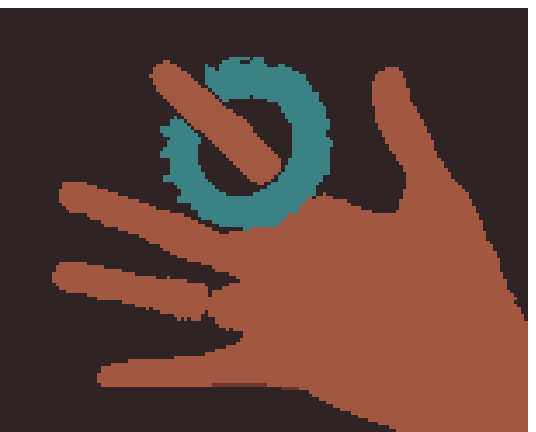}&
\includegraphics[scale=0.6]{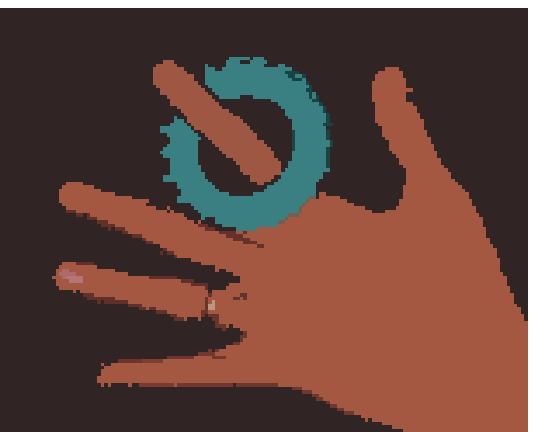}\\
\hspace{-0.2cm}\mbox{a)} & \mbox{b)} & \mbox{c)}\vspace{-0.2cm}
\end{tabular}
\caption{Validation of the iterative selection on the hand image.  a) Original image; b) Non iterative bandwidth selection; c) Iterative bandwidth selection.}
\label{fig:hand1}
\end{center}
\end{figure}

We show other segmentation results on the hand image (Figure \ref{fig:hand1} a)). For the first one 13 clusters are found against 37 for the second. The ring and the nails are not detected by the non iterative method because the selected color bandwidths are too large. The reason is that the regions to be segmented are large, leading to large position bandwidths. Because the bandwidths for position and color are chosen jointly, the color bandwidth are also too large as a result.

\begin{figure}[h!]
\begin{center}
\begin{tabular}{ccc}
\hspace{-0.2cm}\includegraphics[scale=0.6]{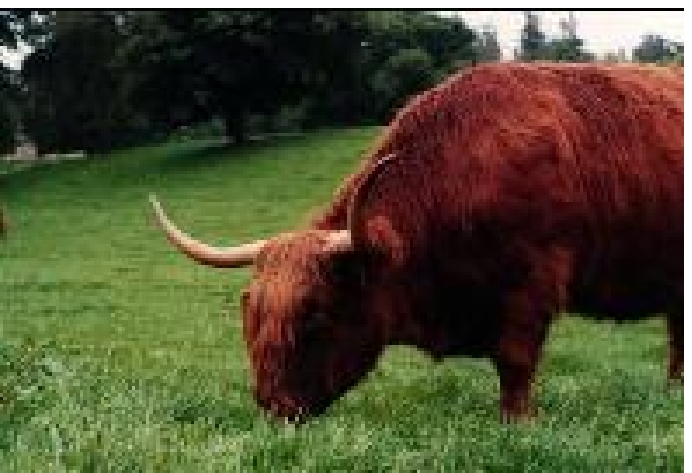}&
\includegraphics[scale=0.6]{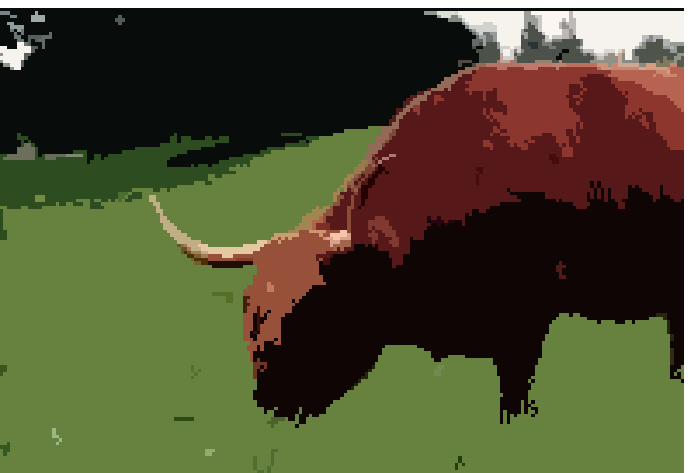}&
\includegraphics[scale=0.6]{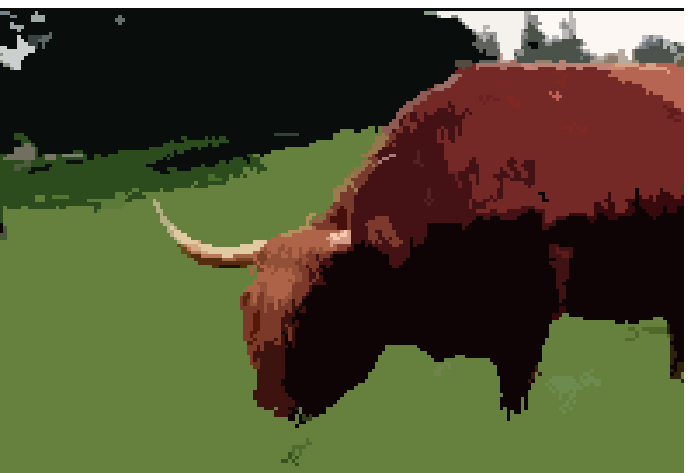}\\
\hspace{-0.2cm}\mbox{a)} & \mbox{b)} & \mbox{c)}\vspace{-0.2cm}
\end{tabular}
\caption{Validation of the iterative selection on the bull image.  a) Original image; b) Non iterative bandwidth selection; c) Iterative bandwidth selection.}
\label{fig:bull1}
\end{center}
\end{figure}

A last result is presented on the bull image (Figure \ref{fig:bull1} a)). Major differences between the segmentations obtained with the two methods are visible on the bull itself. In particular, the iterative algorithm keeps more details on the head of the animal. The number of clusters found by the two algorithms are not so different though. Indeed, the iterative method found 136 clusters while the non-iterative one gave 130 clusters. While more important details are kept on the bull by the iterative method, some little (but less important) clusters are lost on the grass. \\

A surprising result concerns the computational cost. One could think that the non iterative method would be much faster than our iterative algorithm. This is not the case, even sometimes the iterative selection is faster. This can be explained as follows. While the first iterations are run for large bandwidths (mean over all the predefined bandwidths), in subsequent iterations, the best bandwidths have been chosen for the first feature spaces. These bandwidths are more adapted and lead to faster computation of the mean shift partitionings.

To conclude, the iterative method permits to keep more details than the non iterative one, even if the results are, sometimes, visually close. With the iterative method, it is not the same scale that is chosen for all the dimensions. Furthermore, the introduction of our iterative selection method does not cause any computation overhead.

\subsection{Validation of the pseudo balloon mean shift partitioning}
A novelty of our approach is the introduction of the pseudo balloon mean shift partitioning. We compare this partitioning method to the variable mean shift partitioning introduced by Comaniciu in \cite{Comaniciu01}. In \cite{Terrell92} it has been shown that the balloon estimator gives good results when the number of dimensions increases, which led to its use in this paper. The comparison is done as follows. The bandwidth selection using the pseudo balloon mean shift partitioning as described in this paper is first run. Then using the selected bandwidths, the variable bandwidth and the pseudo balloon mean shift partitioning are run and give the final segmentations.

\begin{figure}[h!]
\begin{center}
\begin{tabular}{ccc}
\hspace{-0.2cm}\includegraphics[scale=0.85]{Chalet.eps}&
\includegraphics[scale=0.85]{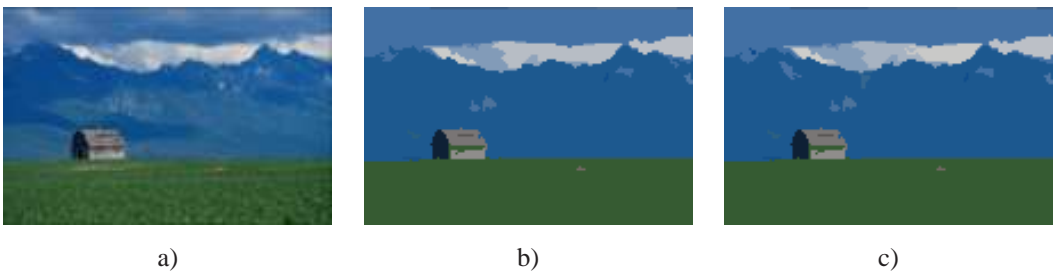}&
\includegraphics[scale=0.85]{Balloon_Balloon_0043_0.eps}\\
\hspace{-0.2cm}\mbox{a)} & \mbox{b)} & \mbox{c)}\vspace{-0.2cm}
\end{tabular}
\caption{Validation of "pseudo balloon mean shift" on the outdoor image. a) Original image; b) Variable mean shift partitioning; c) Pseudo balloon mean shift partitioning.}
\label{fig:chalet2}
\end{center}
\end{figure}

Here again we start by showing the results on the outdoor image. Figure \ref{fig:chalet2} shows the final partitioning for the variable mean shift based on the sample point estimator (b)) and the pseudo balloon mean shift (c)). Nearly the same results were obtained by the two partitioning method. The sample point estimator gave 29 clusters against 31 for the pseudo balloon mean shift partitioning. Few tiny differences can be found in the clouds.

\begin{figure}[h!]
\begin{center}
\begin{tabular}{ccc}
\hspace{-0.2cm}\includegraphics[scale=0.6]{Main.eps}&\includegraphics[scale=0.6]{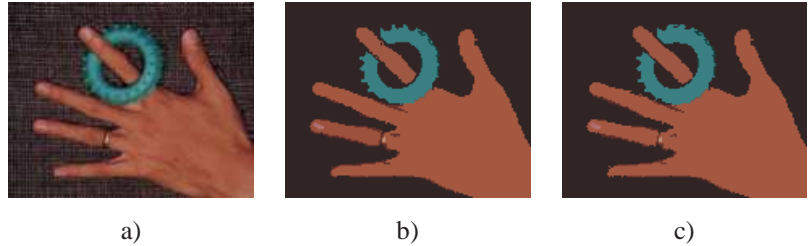}&
\includegraphics[scale=0.6]{Balloon_Balloon_0022_0.eps}\\
\hspace{-0.2cm}\mbox{a)} & \mbox{b)} & \mbox{c)}\vspace{-0.2cm}
\end{tabular}
\caption{Validation of the "pseudo balloon mean shift" on the hand image. a) Original image; b) Variable mean shift partitioning; c) Pseudo balloon mean shift partitioning.}
\label{fig:hand2}
\end{center}
\end{figure}

The next result is on the hand image (Figure \ref{fig:hand2}). The segmentations are again very close with the two estimators. The variable mean shift partitioning \cite{Comaniciu01} gave 35 clusters and the pseudo balloon 37. The segmentation of the forefinger for the variable sample point mean shift is slightly less clean (composed of two clusters).

\begin{figure}[h!]
\begin{center}
\begin{tabular}{ccc}
\hspace{-0.2cm}\includegraphics[scale=0.6]{Taureau.eps}&
\includegraphics[scale=0.6]{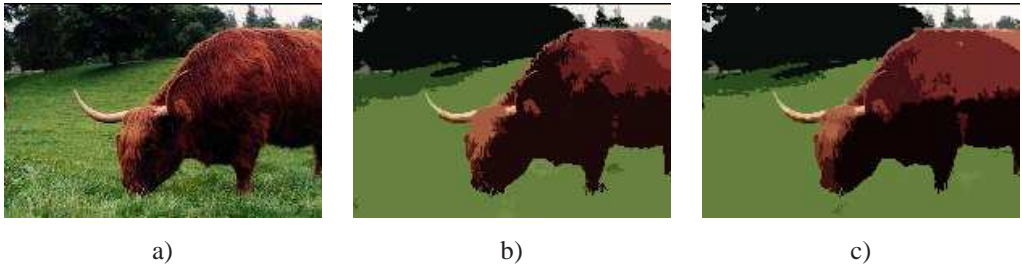}&
\includegraphics[scale=0.6]{Balloon_Balloon_0033_0.eps}\\
\hspace{-0.2cm}\mbox{a)} & \mbox{b)} & \mbox{c)}\vspace{-0.2cm}
\end{tabular}
\caption{Validation of the "pseudo balloon mean shift" on the bull image. a) Original image; b) Variable mean shift partitioning; c) Pseudo balloon mean shift partitioning.}
\label{fig:bull2}
\end{center}
\end{figure}

We end this subsection by presenting the segmentation results on the bull image (Figure \ref{fig:bull2}). Contrary to the two previous results, many differences are visible between the final partitioning obtained with the variable mean shift based on the sample point estimator (Figure \ref{fig:bull2}(b)), which gave 128 clusters, and the pseudo balloon mean shift (c), which led to 136 clusters. The pseudo balloon mean shift permits to keep more details on the head of the bull. Also, the clusters found on the back are less messy than the ones found with the algorithm using the sample point estimator. \\

To conclude, the results presented in this subsection show that the pseudo balloon mean shift partitioning can be as good (or even better) as the variable mean shift of Comaniciu \cite{Comaniciu01}. In addition, the balloon estimator is more adapted to high-dimensional ($d \ge 3$) heterogeneous data ({\it e.g.} five-dimensional data in color segmentation), thanks to the iterative bandwidth selection we introduced in \ref{sec:algobal}.

\subsection{Ordering the feature spaces}
\begin{figure}[h!]
\begin{center}
\begin{tabular}{cc}
\includegraphics[scale=0.8]{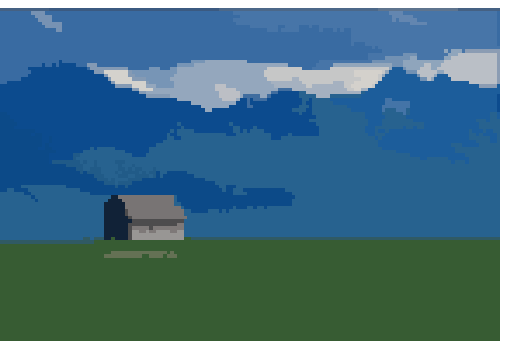}&\includegraphics[scale=0.8]{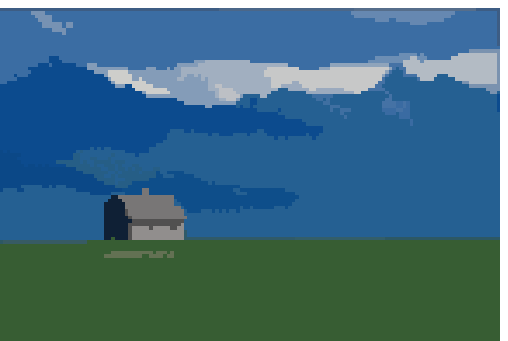}\\
\mbox{a)} & \mbox{b)} \vspace{-0cm}
\end{tabular}
\caption{Ordering the feature spaces. Results of the balloon mean shift partitioning on the outdoor image when the feature spaces are ordered in the following ways: a) x-coordinate, y-coordinate, red channel, green channel, blue channel; b) blue channel, green channel, red channel, y-coordinate, x-coordinate.}
\label{fig:order1}
\end{center}
\end{figure}

\begin{figure}[h!]
\begin{center}
\begin{tabular}{cc}
\includegraphics{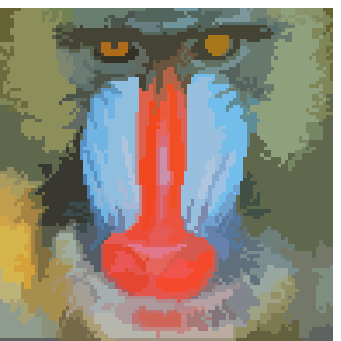}&\includegraphics{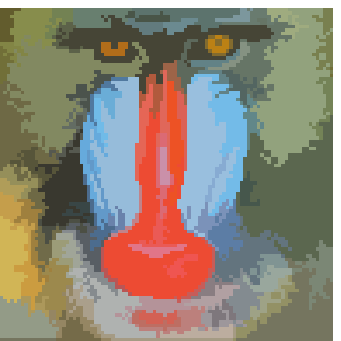}\\
\mbox{a)} & \mbox{b)} \vspace{-0cm}
\end{tabular}
\caption{Ordering the feature spaces. Results of the balloon mean shift partitioning on the baboon image when the feature spaces are ordered in the following ways: a) x-coordinate, y-coordinate, red channel, green channel, blue channel; b) blue channel, green channel, red channel, y-coordinate, x-coordinate.}
\label{fig:order2}
\end{center}
\end{figure}

One could wonder if the order in which the feature spaces are studied is important. In fact it has only a small influence (Figures \ref{fig:order1} and \ref{fig:order2}). These results have been obtained using 9 bandwidths in the range 3-20. 
As judging a segmentation depends on the subsequent application, defining the order in which the feature spaces should be processed is at this stage not really possible. An intuition would be to start with the position before processing successively the feature spaces having the highest noise or the highest contrast in the image.

\section{Conclusion}
Automatic bandwidth selection for kernel bandwidth estimation has become an important research area as the popularity of mean shift methods for image and video segmentation increases. Several methods already exist in literature but none of them is really adapted to the case of multidimensional heterogeneous features. This is the problem we addressed in this paper.
To this end, we first introduced the pseudo balloon mean shift filtering and partitioning to which the kernel bandwidth selection was applied. The convergence of this filtering method has been proved. Following \cite{Comaniciu03a}, the selection is is based on the intuition that a good partition must be stable through scales.
The bandwidth selection method is based on an iterative selection over the different feature subspaces. It allows a richer search of optimal analysis bandwidths than the non iterative method in \cite{Comaniciu03a}. The validity of our algorithm was shown on color image segmentation. Note that our algorithm has also been used for motion detection in \cite{CVPR07}, leading to very promising results. A direction of future research concerns the computation of covariance matrices. It would indeed be valuable to devise a new way of computing them that permits to capture the distribution near cluster's mode and the tails of a cluster.


\bibliographystyle{plain}
\bibliography{biblio}

\appendix
\part*{Appendix \markboth{}{} \addcontentsline{toc}{part}{\bf
Appendix}}
\renewcommand{\theequation}{A.\arabic{equation}}
\setcounter{equation}{0}  
\section*{Proof of convergence of the pseudo balloon mean shift filtering}  
\label{ap:balloon1}
The balloon kernel density estimator is defined as:
\begin{equation}
\widehat f({\bf x}) = \frac{c_k}{n}\sum_{i=1}^n \frac{1}{|{\bf H}({\bf x})|^{1/2}} k(\|{\bf H}({\bf x})^{-1/2}({\bf x}-{\bf x}^{(i)})\|^2)~~.
\end{equation}
The proof of convergence of mean shift filtering using this estimator is closed to the one of the fixed bandwidth mean shift filtering \cite{Comaniciu02}. We first show that {\small $\widehat f$} is convergent for the trajectory points defined in algorithm \ref{algo:ms}, {\it i.e.} that {\small$\widehat f({\bf y}^{(j)})$} converges when $j$ becomes large if $m({\bf x})$ is defined as in (\ref{eq:bms2}). Since {\small $n$} is finite, {\small$\widehat f$} is bounded : {\small $ 0 < \widehat f(x) \le \frac{c_k}{n|{\bf H}({\bf x})|^{1/2}}  $}. It is then sufficient to show that {\small $\widehat f$} is strictly increasing or decreasing. Since the bandwidth {\small ${\bf H}({\bf x})$} is constant for all trajectory points {\small ${\bf y}^{(j)}$} associated to the estimation point {\small ${\bf x}$}, we get:
\begin{equation}
\begin{split}
    \widehat f({\bf y}^{(j+1)})&-\widehat f({\bf y}^{(j)}) \\
    &= \frac{c_k}{n|{\bf H}({\bf x})|^{1/2}}  \sum_{i=1}^{n}\Big(k(\|{\bf H}({\bf x})^{-1/2}({\bf y}^{(j+1)}-{\bf x}^{(i)})\|^2)-  k(\|{\bf H}({\bf x})^{-1/2}({\bf y}^{(j)}-{\bf x}^{(i)})\|^2)\Big)  ~~.
\end{split}
\end{equation}
The convexity of the profile {\small $k$} implies that:
 \begin{equation*}
    \forall ({x}_1,{x}_2) \in [0,+\infty)   ~~k({x}_2) \ge k({x}_1) + k'({x}_1)({x}_2-{x}_1)~~ ,
  \end{equation*}
and thus:
\begin{equation}
  \begin{split}
   \widehat f({\bf y}^{(j+1)})- \widehat f({\bf y}^{(j)})\ge \frac{c_k}{n|{\bf H}({\bf x})|^{1/2}} & \sum_{i=1}^{n} k'(\|{\bf H}({\bf x})^{-1/2}({\bf y}^{(j)}-{\bf x}^{(i)})\|^2)\\
&\Big(\|{\bf H}({\bf x})^{-1/2}({\bf y}^{(j+1)}-{\bf x}^{(i)})\|^2- \|{\bf H}({\bf x})^{-1/2}({\bf y}^{(j)}-{\bf x}^{(i)})\|^2 \Big) .
 \end{split}
\end{equation}
We assume that {\small${\bf H}({\bf x})^T ={\bf H}({\bf x})$}. Developing the last term and using the definition of the mean shift vector (equation \ref{eq:bms2}) implies after some manipulations:
\begin{equation}
  \begin{split}
    \widehat f({\bf y}^{(j+1)})&- \widehat f({\bf y}^{(j)})\\
 &\ge \frac{c_k}{n|{\bf H}({\bf x})|^{1/2}}  \sum_{i=1}^{n} k'(\|{\bf H}({\bf x})^{-1/2}({\bf y}^{(j)}-{\bf x}^{(i)})\|^2)\Big(\|{\bf H}({\bf x})^{-1/2}({\bf y}^{(j+1)}-{\bf y}^{(j)})\|^2\Big) .
  \end{split}
\end{equation}
Summing terms of previous equation for index {\small $j,j+1,\ldots,j+m-1$}, and introducing\\
 {\small $M=\argmin_{l \ge 0}k(\|{\bf H}({\bf x})^{-1/2}({\bf y}^{(l)}-{\bf x}^{(i)})\|^2$}  results in:
\begin{equation}\label{eq:proof1}
  \widehat f({\bf y}^{(j+m)})-\widehat f({\bf y}^{(j)}) \ge \frac{c_k}{n|{\bf H}({\bf x})|^{1/2}} M \|{\bf H}({\bf x})^{-1/2}({\bf y}^{(j+m)}-{\bf y}^{(j)})\|^2 \ge 0.
\end{equation}
We have shown that the sequence {\small $\{\widehat f({\bf y}^{(j)})\}_{j=1,2\ldots}$} is strictly increasing, bounded, and thus convergent. It is also a Cauchy sequence. Inequality (\ref{eq:proof1}) implies that {\small $\{{\bf y}^{(j)}\}_{j=1,2\ldots}$} is also a Cauchy sequence with respect to Mahalanobis norm, hence with respect to Euclidean norm (by vertue of norm equivalence in $ \mathbb{R}^d$). This proves the convergence of trajectory points towards a local mode of ${\widehat f}$.




\end{document}